# Cyber Physical System Information Collection: Robot Location and Navigation Method Based on QR Code

Hongwei Li[1] · Tao Xiong[2]

**Abstract**

In this paper, we propose a method to estimate the exact location of a camera in a cyber-physical system using the exact geographic coordinates of four feature points stored in QR codes(Quick response codes) and the pixel coordinates of four feature points analyzed from the QR code images taken by the camera. Firstly, the P4P(Perspective 4 Points) algorithm is designed to uniquely determine the initial pose estimation value of the QR coordinate system relative to the camera coordinate system by using the four feature points of the selected QR code. In the second step, the manifold gradient optimization algorithm is designed. The rotation matrix and displacement vector are taken as the initial values of iteration, and the iterative optimization is carried out to improve the positioning accuracy and obtain the rotation matrix and displacement vector with higher accuracy. The third step is to convert the pose of the QR coordinate system with respect to the camera coordinate system to the pose of the AGV(Automated Guided Vehicle) with respect to the world coordinate system. Finally, the performance of manifold gradient optimization algorithm and P4P analytical algorithm are simulated and compared under the same conditions.One can see that the performance of the manifold gradient optimization algorithm proposed in this paper is much better than that of the P4P analytic algorithm when the signal-to-noise ratio is small.With the increase of the signal-to-noise ratio,the performance of the P4P analytic algorithm approaches that of the manifold gradient optimization algorithm.when the noise is same,the performance of manifold gradient optimization algorithm is better when there are more feature points.

This research was partially supported by the Special Project for Key Fields of Ordinary Universities of Guangdong Provincial Department of Education (2022ZDZX1057), Characteristic Innovation Project for Ordinary Universities of Guangdong Provincial Department of Education (2023KTSCX267), Basic and Applied Basic Research Project of Guangzhou Basic Research Program (1717).

Hongwei Li    e-mail:mclhw@gdcp.edu.cn

Tao Xiong     e-mail:290418228@qq.com

1    Maritime College in Guangdong Communication Polytechic, GuangZhou, China ,510650

2    Guangzhou Haige Communications Industry Group Co.，Ltd，Guangzhou, China ,510663

## 1. Introduction

CPS(cyber physical systems) is a new field of research, first proposed by the National Science Foundation in 2006. At present, the concept of the parties to the CPS, definition is not the same, but in general, the nature of the CPS is to build a set of information space and physical space based on the data flow between the state of perception, real-time analysis, scientific decision-making, accurate implementation of closed loop can assign system, solve the production, the complexity and uncertainty problems in the process of application service, Improve the efficiency of resource allocation and realize resource optimization[1]. State perception is to perceive the running state of the material world through a variety of sensors[2]. Real-time analysis is the transformation of data, information and knowledge through industrial software. Scientific decision-making is to realize data flow and knowledge sharing in heterogeneous systems through big data platform[3]. Precise execution is to realize the feedback response to the decision by mechanical hardware such as controller and actuator[4]. However, all this depends on a real-time, reliable and secure network.

Information physical system includes the ubiquitous environment awareness, embedded computing, network communication and network control system engineering in the future, so that the physical system has the functions of computing, communication, precise control, remote cooperation and autonomy[5]. It pays attention to the close combination and coordination of computing resources and physical resources, and is mainly used in some intelligent systems such as device interconnection, Internet of things sensing, smart home, robots, intelligent navigation, etc[6].

Intelligent mobile robot is a kind of robot system that can sense the environment and its own state through sensors, and realize target-oriented autonomous movement (called navigation) in the environment with obstacles, so as to complete certain operation functions. Mobile robots can be used to complete many important tasks in industry, agriculture, medical, service and other industries, and navigation technology is the core of mobile robot research. Among all kinds of navigation technologies, visual navigation is regarded as an important development direction of AGV navigation technology in the future due to its advantages of wide detection range, complete target information and low cost.At present, two-dimensional code is widely used in many fields[7], depending on its independent generation of information, independent on database, low cost and easy production. In addition, it can be moved with carrier. QR code is a typical matrix two-dimensional code with the characteristic structure function graph. As long as the version specification is determined, its function graph number and the relative position are fixed. So, QR code can be introduced into the visual navigation.

There have been many studies on using two-dimensional codes to replace traditional natural feature targets for robot visual navigation.The work [8] stores the coordinate information of the spatial location of the QR code image in the QR code,similar to the magnetic nail navigation method, the two-dimensional code is laid at intervals or at specific locations on the AGV moving path.In the process of AGV moving, the two-dimensional code is scanned by the vehicle borne industrial camera.The position and attitude information of AGV is determined by identifying the current two-dimensional code, and the Angle error and position error of AGV on the course are calculated, so as to correct the movement route on AGV and drive to the next two-dimensional code with more correct attitude.This method is simple and easy to implement, but the positioning accuracy is low.The work[9] designs a navigation and positioning method that uses a matrix two-dimensional code as an absolute position label to assist in correcting the mileage and the cumulative error generated in the inertial measurement system navigation process.The two methods[8,9] use the location information stored in QR code to locate, which has low positioning accuracy and strict restrictions on AGV's travel route.

The work[10] designs a visual navigation based on QR Code, but it is only a visual navigation based on the functional graphics and coding regions in different QR code pictures.The work[11] proposes a composite navigation method based on QR code visual positioning,ground grid map is constructed by using the discrete QR code landmark．The relative position between landmark and robot in the grid map is calculated by vision positioning QR code， including edge extraction， line detection，rectangle feature detection，recognition of QR code feature and deviation calculation. The work[12] determines the center coordinates of the rectangular area by extracting the rectangular outline of the two-dimensional code in the acquired image; then, the position information was obtained through coordinate rotation and transformation of the unified coordinate system.The above methods calculate the camera position by extracting the graphic contour of the QR code and using the spatial relative position information of the camera and the QR code, so the positioning accuracy is low.When the camera takes the QR code picture, the position requirements of the camera are very strict, and the motion trajectory robustness of the camera is poor.

The work[13] uses the location information stored in the DM code and the salient feature l edge of the DM code for visual positioning, and calculates the AGV position through coordinate transformation. This method is similar to the method described in this paper, but the detailed design steps are not given in this paper.The work[14] presents a method of visual measurement based on QR code to realize the navigation and positioning of agricultural robot in closed-type orchard,this method boasts good real-time performance as the longest consuming time of applying this method to process an image for visual measurement is 0.23ms.in addition,the repeatability accuracy tests show that the maximum translation deviation is less than 3.2cm and the maximum rotation deviation is less than 2.9°in the range of measurement. This paper successfully uses P4P algorithm to realize AGV localization.[13,14] these two methods use the location information stored in the QR code and the pixel coordinates of the QR code feature points resolved in the QR code image taken by the camera for positioning.

The visual navigation and positioning method based on QR code has been widely used in indoor or outdoor environment.The work[15]proposes using a fusion algorithm between the IMU and PnP solution for an automatic landing assistant system for landing a fixed-wing in outdoor

environment.The work[16]proposes a QR code image processing mechanism for library robot positioning and navigation system based on the visual technology. The work[17,18]proposed an approach using QR codes as landmarks for localization and navigation for a mobile robot in indoor environment.Of course, in order to meet the different use needs of indoor and outdoor environments, flexible design can be carried out based on their respective advantages of visual navigation and QR code recognition. For example, in order to improve positioning accuracy, some studies have designed binocular visual positioning and navigation technology based on QR code recognition[19]. In order to achieve light weight, low power consumption and small physical size, some researches use smart phones and QR codes to achieve positioning[20].

The above research shows that two-dimensional code based visual navigation and positioning has been widely used in various workplaces. However, these papers only introduced the system implementation or working methods, but did not explain the navigation and positioning algorithms in detail. Even if it is introduced, it only lists the implementation steps of the P4P algorithm and has no practical reference value for other researchers. In order to provide practical guidance for researchers, this paper describes the PnP algorithm process based on two-dimensional code localization in detail. This paper not only introduces the key parameters of PnP algorithm, such as the rotation matrix, position calculation and attitude Angle calculation method, but also designs the gradient optimization algorithm to optimize the P4P algorithm iteratively, which effectively improves the positioning accuracy. The simulation results show that the gradient optimization algorithm has obvious optimization effect on the P4P algorithm. At the same time, this paper puts forward the precautions of using the algorithm described in this paper and suggestions for further improving the positioning accuracy. One of the measures is to obtain higher positioning accuracy by increasing the number of feature points on the two-dimensional code for calculation. The more the number of feature points, the higher the positioning accuracy of P4P algorithm is verified by simulation. In this case, manifold gradient optimization algorithm can also improve the performance of P4P algorithm.

## 2. Description of camera pose estimation method using QR code

A rigid body has six degrees of freedom in three-dimensional space: the coordinates of the origin of its fixed coordinate system in the world coordinate system and three attitude angles.The fixed coordinate system means that the coordinate system is fixed on the rigid body, and its relative position with the rigid body does not change. These six degrees of freedom are called the pose of the rigid body. The meaning of pose estimation is to calculate all the six parameters.

Since QR code has the function of storing information, QR can store the actual coordinate information of its four feature points in the process of positioning.When the QR code is captured by the industrial camera installed on AGV, the actual coordinates of the four feature points of the QR code can be analyzed. In addition, the pixel coordinates of the four feature points in the image can be obtained from the QR code picture taken by the camera.the actual coordinates of the four feature points stored in the QR code and the pixel coordinates of the four feature points in the QR image can be calculated to obtain the camera's pose relative to the QR code coordinate system.Since the positional relationship between the QR code coordinate system and the world coordinate system is known, the pose of the camera relative to the world coordinate system can be obtained. The camera is installed on the AGV vehicle through the frame, and is fixedly connected with the wheeled mobile robot, so the camera and the wheeled mobile robot can be regarded as a rigid body. The pose of the camera relative to the world coordinate system is the actual spatial position of the AGV trolley.

The algorithm is divided into three steps. In the first step, the P4P algorithm is designed, which uses the selected four feature points of QR code to uniquely determine the initial value of the QR coordinate system relative to the camera coordinate system (the position is represented by rotation matrix and displacement vector, with low numerical accuracy); In the second step, the gradient optimization algorithm is designed, and the rotation matrix and displacement vector are used as the initial values for iterative optimization; In the third step, the position and attitude of the QR coordinate system relative to the camera coordinate system (the optimized rotation matrix and displacement vector) is transformed into the position and attitude of AGV relative to the world coordinate system.

## 3. Principle of P4P algorithm for camera pose estimation

Establish a rigid body fixed-link coordinate system $oxyz$, the origin can be arbitrarily selected in the rigid body or its extended part, which can be any point in space. Generally, some special points will be selected for the convenience of calculation, select the focal point in the camera as the coordinate origin, and select four characteristic points A, B, C, D on the QR code plane. The movement in Euclidean space can be decomposed into two parts: the translation of the origin and the rotation of the coordinate system[21], where the rotation has nothing to do with the selection of the coordinate origin.

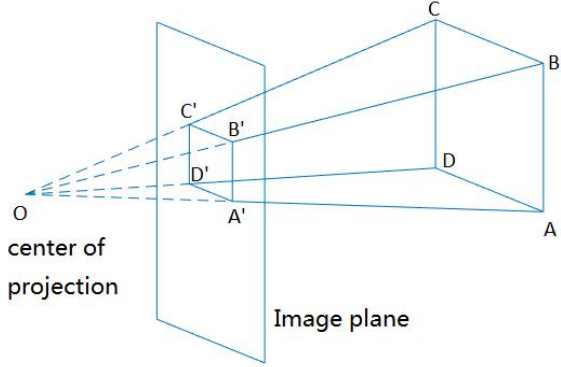

**Figure 1:** P4P algorithm diagram

Assuming that the coordinate system of the rigid body coincides with the world coordinate system, the coordinate change of any point $P$ on the rigid body in the world coordinate system is:

$$\begin{pmatrix} x' \\ y' \\ z' \end{pmatrix} = R \begin{pmatrix} x \\ y \\ z \end{pmatrix} + t \qquad (1)$$

$R$ is the rotation matrix, which is a third-order orthogonal matrix, and $t$ is the displacement vector.

Take the camera coordinate system as the reference coordinate system, and the QR code as a rigid body. Then the relationship between the coordinates of any point on the QR code coordinate system and the pixel coordinates of this point projected on the camera plane is:

$$z' \begin{pmatrix} u' \\ v' \\ 1' \end{pmatrix} = K \begin{pmatrix} x' \\ y' \\ z' \end{pmatrix} \qquad (2)$$

From (1)(2),

$$z' \begin{pmatrix} u' \\ v' \\ 1' \end{pmatrix} = K \begin{pmatrix} x' \\ y' \\ z' \end{pmatrix} = K \left( R \begin{pmatrix} x \\ y \\ z \end{pmatrix} + t \right) \qquad (3)$$

$K$ is the internal parameter matrix after camera calibration, let the feature point plane be the $oxy$ plane of this point's fixed-link coordinate system, the coordinates of the four feature points A, B, C, and D on the QR code plane in their fixed-link coordinate system are respectively

$\begin{pmatrix} x_1 \\ y_1 \\ 0 \end{pmatrix}$、$\begin{pmatrix} x_2 \\ y_2 \\ 0 \end{pmatrix}$、$\begin{pmatrix} x_3 \\ y_3 \\ 0 \end{pmatrix}$、$\begin{pmatrix} x_4 \\ y_4 \\ 0 \end{pmatrix}$ ,Substituting the four values into equation (3), one can get the following results:

$$z_i' \begin{pmatrix} u_i' \\ v_i' \\ 1' \end{pmatrix} = K \begin{pmatrix} x_i' \\ y_i' \\ z_i' \end{pmatrix} = K(r_1 \ r_2 \ t) \begin{pmatrix} x_i \\ y_i \\ 1 \end{pmatrix} \equiv H \begin{pmatrix} x_i \\ y_i \\ 1 \end{pmatrix} \qquad (4)$$

$R = (r_1 \ r_2 \ r_3), i = 1, 2, 3, 4$. For the equation system (4), the matrix $H$ can be solved by a constant factor $s$, and

$$(r_1 \ r_2 \ t) = K^{-1} H \qquad (5)$$

$$r_3 = r_1 \times r_2 \qquad (6)$$

Therefore, the complete pose $R$ and $t$ can be solved. Equation (6) ensures that the rotation is carried out in the same coordinate system., set the camera coordinate system to the right-hand coordinate system., then this equation can ensure the z-axis direction of the feature point plane. At the same time, by setting the constant factor s to be positive, the solution that conforms to equation (4) and makes the calculated feature point depth negative is eliminated.

Therefore, the P4P algorithm uniquely determines the transformation result between the right-hand coordinate system of the feature plane and the camera right-hand coordinate system, that is, the pose.

The following describes the calculation method of the matrix $H$.

One can define some matrices as:

$$q_1 = \begin{pmatrix} x_1 \\ y_1 \\ 1 \end{pmatrix}, q_2 = \begin{pmatrix} x_2 \\ y_2 \\ 1 \end{pmatrix}, q_3 = \begin{pmatrix} x_3 \\ y_3 \\ 1 \end{pmatrix}, q_4 = \begin{pmatrix} x_4 \\ y_4 \\ 1 \end{pmatrix},$$

$$Q = (q_1 \ q_2 \ q_3),$$

$$p_1 = \begin{pmatrix} x_1' \\ y_1' \\ 1 \end{pmatrix}, p_2 = \begin{pmatrix} x_2' \\ y_2' \\ 1 \end{pmatrix}, p_3 = \begin{pmatrix} x_3' \\ y_3' \\ 1 \end{pmatrix}, p_4 = \begin{pmatrix} x_4' \\ y_4' \\ 1 \end{pmatrix},$$

$$P = (p_1 \ p_2 \ p_3).$$

$$v = P^{-1} p_4 \tag{7}$$

$$r = Q^{-1} q_4 \tag{8}$$

The matrices $P$ and $Q$ are both invertible, because any three of the feature points $q_i$ and any three of the corresponding pixel points $p_i$ are not collinear. The correspondence between 1, 2, 3, 4 and A, B, C, D can be changed arbitrarily. In some cases, when the plane of the feature points is torsion large, and then subject to noise interference, three of the pixel points $p_i$ may be close to collinear, which seriously affects the performance. At this time, it doesn't make sense to choose three non-collinear ones. The four characteristic points of the QR code are rectangular, and the size is not large. If the three points are collinear, the fourth point is not far away.

The coordinates of the feature point $q_i$ are measured in advance and will not change due to the change of the scene in practice, unless the QR code is distorted by the environment, but this is a maintenance problem. The positioning algorithm can not obtain this information. Even if the image processing algorithm in the previous stage has done distortion and distortion correction, which has nothing to do with the positioning algorithm.

Define matrix $U$ as $U = \begin{pmatrix} z_1' & 0 & 0 \\ 0 & z_2' & 0 \\ 0 & 0 & z_3' \end{pmatrix}$, by (4) one can obtain:

$$HQ = PU \Leftrightarrow H = PUQ^{-1} \tag{9}$$

$$z_4' p_4 = H q_4 \tag{10}$$

Substitute (9) into (10), one can obtain:

$$z_4' p_4 = PUQ^{-1} q_4 \Leftrightarrow z_4' P^{-1} p_4 = UQ^{-1} q_4 \tag{11}$$

Substitute (7) and (8) into (10),

$$z_4' v = Ur \tag{12}$$

Expand the formula to get:

$$z_4' v_1 = z_1' r_1, z_4' v_3 = z_3' r_3 \Rightarrow \frac{z_1'}{z_3'} = \frac{r_3}{r_1} \frac{v_1}{v_3} \tag{13}$$

$$z_4' v_2 = z_2' r_2, z_4' v_3 = z_3' r_3 \Rightarrow \frac{z_2'}{z_3'} = \frac{r_3}{r_2} \frac{v_2}{v_3} \tag{14}$$

In the two equations (13) and (14), the denominator is the depth of the feature point, which is a positive number at least not less than the focal length, so there is no need to worry about being divided by zero.

$$\text{define } W = \frac{1}{z_3'} U = \begin{pmatrix} \frac{z_1'}{z_3'} & 0 & 0 \\ 0 & \frac{z_2'}{z_3'} & 0 \\ 0 & 0 & 1 \end{pmatrix} = \begin{pmatrix} \frac{r_3}{r_1} \frac{v_1}{v_3} & 0 & 0 \\ 0 & \frac{r_3}{r_2} \frac{v_2}{v_3} & 0 \\ 0 & 0 & 1 \end{pmatrix}$$

Substituting the matrix $W$ into (9), one can get:

$$H = \frac{1}{z_3'} PWQ^{-1} \equiv \frac{1}{z_3'} T \equiv sT \tag{15}$$

In equation (15), the three matrices on the right can be calculated, and the product $T$ differs from the required matrix $H$ by a constant multiplier factor $\frac{1}{z_3'}$, which is the reciprocal of the axial distance (depth) of the characteristic point 3 from the optical center, which is Unknown.

But there are constraints: $R$ is an orthogonal matrix. Assuming that the matrix $T$ has been calculated, combining (4), one can obtain:

$$(r_1 \ r_2 \ t) = sK^{-1}T = (sg_1 \ sg_2 \ sg_3) \tag{16}$$

Now it is required to take a reasonable positive number $s$ (excluding unreasonable negative solutions), so that define $sg_1$, $sg_2$ are orthonormal vectors, this goal is impossible to achieve accurately.. List the smallest solution in the least square sense[22-24]:

$$s_o = \arg\min_s [(s|g_1|-1)^2 + (s|g_2|-1)^2] \tag{17}$$

one can define $|g_i| = \sqrt{g_i^H g_i}$,

So one can obtain $S_o = \frac{|g_1| + |g_2|}{|g_1|^2 + |g_2|^2}$ (18)

$r_1 = \frac{|g_1| + |g_2|}{|g_1|^2 + |g_2|^2} g_1$, (19)

$r_2 = \frac{|g_1| + |g_2|}{|g_1|^2 + |g_2|^2} g_2$, (19)

$t = \frac{|g_1| + |g_2|}{|g_1|^2 + |g_2|^2} g_3$, (20)

$r_3 = r_1 \times r_2$, (21)

So far, the P4P algorithm is completed, and the rotation matrix $R$ and translation vector $t$ are obtained.

But the matter is not finished, the rotation matrix $R$ is not completely orthogonal, and it is not on the 3D manifold and Lie group so (3). The following gradient algorithm needs the initial value on so (3), so $R$ must be orthogonalized and transferred to manifold. The orthogonalization is carried out by gramm Schmidt algorithm[25,26], which is not introduced here.

## 4. Manifold gradient optimization algorithm

By giving the coordinates of some characteristic points on the plane of QR code, $R$ and $t$ are calculated, so that the pixel coordinates of these points and the actual pixel coordinates are in good agreement. When four points are given, the pose can be solved uniquely. In general, the objective function can be set up:

$F(R, t) = \frac{1}{2} f(R, t)^T f(R, t) = \frac{1}{2} \sum_{i=1}^{n} |p_i - p_i'|^2$ (22)

or $F(R, t) = \frac{1}{2} f(R, t)^T f(R, t) = \frac{1}{2} \sum_{i=1}^{n} |z_i' p_i - z_i p_i'|^2$ (23)

Find suitable $R$ and $t$ to make the function $F$ as small as possible. define $p_i = \begin{pmatrix} u_i \\ v_i \end{pmatrix}$, $p_i' = \begin{pmatrix} u_i' \\ v_i' \end{pmatrix}$, $p_i$ is the actual pixel coordinates of the feature point, $p_i'$ is the pixel coordinate, which is calculated from the coordinates of the feature point on the QR code plane and the parameters $R$ and $t$. $n$ is the number of feature points. The actual pixel coordinates of the feature points are obtained from the picture by the method of image processing. The matrix $K$ is the camera parameter matrix, which is a third-order square matrix.

Write it as a row matrix $K = \begin{pmatrix} k_1 \\ k_2 \\ k_3 \end{pmatrix}$, both $z_i'$ and $p_i'$ are functions of $R$ and $t$.

In the optimization process of equation (22), the derivative of $\frac{1}{z'}$ is needed. The form of its derivative makes the convergence slow in the larger region of $z'$, and the convergence speed is inconsistent in three coordinate directions. Therefore, equation (23) is adopted to optimize the objective function. The characteristic of equation (23) is that the farther the characteristic point is, the larger the proportion of the error will be. Equation (23) is a quadratic function of position and a complex nonlinear function of attitude, and its extreme value characteristics are relatively complex.

Equation (23) has at least two global minimum points, one is the correct pose point $R_0, t_0$, the other is $R_0', -t_0$, where the first two columns of $R_0'$ are the first two columns of $R_0$ multiplied by $-1$, and the third column is the third of $R_0$ Column. This kind of minimum point is eliminated by the judgment of the depth sign, but otherwise, the algorithm may fall into a local minimum point. Therefore, appropriate initial values must be provided for the numerical optimization algorithm.

In order to solve the optimization problem of equation (23), the value of the orthogonal matrix $R$ needs to be optimized. The number of elements of the three-dimensional orthogonal matrix is 9, which are located in the 9-dimensional Euclidean space, but the set of three-dimensional orthogonal matrix forms a three-dimensional manifold, which is also a Lie group. At present, the PnP problem mainly focuses on two research directions:1. Using least squares in Euclidean space to find the optimization, and then using a certain method to "migrate" it to a "appropriate" point on the manifold; 2. The least squares optimization in Euclidean space with constraint conditions is used, and the constraint condition is matrix orthogonality; 3. Find the optimization on the manifold.

Method 1 is found in some PnP problems. The main problem lies in the process of "migration". There is no effective method to solve it, and the accuracy is usually poor. For Method 2, the optimization problem with constraints is usually very troublesome, especially when the constraints are very difficult to simplify, such as the orthogonality constraints of the rotation matrix. Therefore, use the third method to optimize.

People have not mastered the technology of numerical optimization on the manifold, but with the help of the local isomorphism relationship between Lie groups and Lie algebras, the optimization on the manifold is transformed into the optimization in tangent space. Thus, the well-known optimization method in linear space can be used to optimize the target value. In any gradient direction of the tangent space, the parameter value is in the space, so the required matrix is on the manifold.

L-M method is used to optimize the numerical gradient. Different from the optimization in Euclidean space, the independent variables are mapped to Li algebra. L-M method relies on error vector function $f(R,t)$ and Jacobi matrix[27].

In multivariable calculus, the derivative is expressed by matrix, and the elements in row $i$ and column $j$ correspond to the derivative of the i-th component of vector function to the j-th independent variable. According to this, the chain rule of derivation can be written as a series of matrix multiplication., the order of the product corresponds to the writing of the chain rule. The derivative matrix is called Jacobi matrix. In this paper, Jacobi matrix $J$ is a matrix of order $2n \times 6$, where $N$ is the number of characteristic points.

One can set the independent variable $\boldsymbol{\xi} = \begin{pmatrix} \boldsymbol{s} \\ \boldsymbol{\rho} \end{pmatrix}$,

where $\boldsymbol{s} = \begin{pmatrix} x \\ y \\ z \end{pmatrix}$, $\boldsymbol{s}$ is the coordinate, $\boldsymbol{\rho} = \begin{pmatrix} \rho_1 \\ \rho_2 \\ \rho_3 \end{pmatrix}$, $\boldsymbol{\rho}$ is the point in the Lie algebraic space corresponding to the rotation matrix $\boldsymbol{R}$, then

$$f = \begin{pmatrix} z_1' u_1' - z_1' u_1 \\ z_1' v_1' - z_1' v_1 \\ z_2' u_2' - z_2' u_2 \\ z_2' v_2' - z_2' v_2 \\ \cdot \\ \cdot \\ z_i' u_i' - z_i' u_i \\ z_i' v_i' - z_i' v_i \\ \cdot \\ \cdot \\ z_N' u_N' - z_N' u_N \\ z_N' v_N' - z_N' v_N \end{pmatrix} \quad (24)$$

One can set $\boldsymbol{f}^i = \begin{pmatrix} z_i' u_i' - z_i' u_i \\ z_i' v_i' - z_i' v_i \end{pmatrix}$

$$\boldsymbol{J} = \frac{\partial \boldsymbol{f}}{\partial \boldsymbol{\xi}} = \begin{pmatrix} \frac{\partial \boldsymbol{f}^1}{\partial \boldsymbol{\xi}} \\ \frac{\partial \boldsymbol{f}^2}{\partial \boldsymbol{\xi}} \\ \cdot \\ \cdot \\ \frac{\partial \boldsymbol{f}^i}{\partial \boldsymbol{\xi}} \\ \cdot \\ \cdot \\ \frac{\partial \boldsymbol{f}^N}{\partial \boldsymbol{\xi}} \end{pmatrix} \quad (25)$$

$$\frac{\partial \boldsymbol{f}^i}{\partial \boldsymbol{\xi}} = \frac{\partial \boldsymbol{f}^i}{\partial \boldsymbol{w}^i} \frac{\partial \boldsymbol{w}^i}{\partial \boldsymbol{\xi}} = \begin{pmatrix} k_{1,1} & k_{1,2} & k_{1,3} - u_i \\ k_{2,1} & k_{2,2} & k_{2,3} - v_i \end{pmatrix} \frac{\partial \boldsymbol{w}^i}{\partial \boldsymbol{\xi}} \quad (26)$$

where $w^i = \begin{pmatrix} x_i' \\ y_i' \\ z_i' \end{pmatrix}$,

The mapping from Lie algebra to Lie group is exponential mapping. In the optimization process, the increment of Lie algebra is calculated and then mapped back to Lie group. Since the multiplication of three-dimensional orthogonal matrix is not commutative, the addition without Lie algebra corresponds to the relationship of Lie group multiplication . Therefore, the left multiplication disturbance model is used for derivative calculation. The form of the derivative matrix is as follows[28]:

$$\frac{\partial w^i}{\partial \xi} = (I - (Rs + t)) \qquad (27)$$

Write the optimized parameters $R$ and $t$ in matrix form:

$T = \begin{pmatrix} R & t \\ 0^T & 1 \end{pmatrix}$. In the process of parameter optimization by numerical algorithm, the results $R$ and $t$ of P4P algorithm are taken as the initial values of iteration, and the parameter increment will be given in each iteration. When the left multiplication model is used for operation, it is necessary to map the parameter increment $\Delta\xi = \begin{pmatrix} \Delta s \\ \Delta \rho \end{pmatrix}$ on Lie algebra into matrix $\Delta T$ and multiply it with $T$ to obtain the updated matrix $T' = \Delta T T$,

$$\Delta T = \begin{pmatrix} \Delta R & V\Delta s \\ 0^T & 1 \end{pmatrix}, \qquad (28)$$

set $\theta = |\Delta\rho|$, then $\Delta\rho = \theta a$.

$$\Delta R = I\cos\theta + (1-\cos\theta)aa^T + a^\wedge \sin\theta \qquad (29)$$

$$V = \frac{\sin\theta}{\theta}I + \left(1 - \frac{\sin\theta}{\theta}\right)aa^T + \left(1 - \frac{\cos\theta}{\theta}\right)a^\wedge \qquad (30)$$

The above algorithm realizes the numerical optimization of rotation matrix $R$ and displacement vector $t$ on manifolds.

# 5. Determination of attitude angle of AGV trolley

## 5.1. Calculation of rotation matrix and position calculation

The final goal of this paper is to get the position and attitude of the AGV relative to the world coordinate system. The position and attitude of QR code coordinate system relative to camera coordinate system is calculated by the above algorithm, so it needs to be transformed.

The pose of coordinate system A relative to coordinate system B is $R, t$, which refers to any point in space, if its coordinate in A is $a$ and its coordinate in B is $b$, then there is the following relationship: $b = Ra + t$. Among them, $R$ is the rotation matrix, that is, the transformation matrix of the vector in the coordinate system A to the coordinate system B. In practice, it can be recorded as $R_A^B$, and $R_A^B$ is the orthogonal matrix. $R_B^A = (R_A^B)^{-1} = (R_A^B)^H$, $t$ can be recorded as $t_A^B$, which is the subscript the coordinates of the origin of the indicated coordinate system in the coordinate system indicated by the superscript.

For a point $P$ in space, suppose its coordinate in the QR code coordinate system is $p_o$, the coordinate in the camera coordinate system is $p_c$, the coordinate in the AGV coordinate system is $p_v$, and the coordinate in the world coordinate system is $p_w$. Suppose the pose of the QR code coordinate system relative to the world coordinate system is $R_o^w, t_o^w$, and the pose of the AGV coordinate system relative to the camera coordinate system is $R_v^c, t_v^c$. Then according to the algorithm in this article, the symbol $R$ that has been used before is changed to $R_o^c$, $t$ Recorded as $t_o^c$. In this way, the order of transformation between coordinate systems is clear at a glance. Then there is

$$p_c = R_o^c p_o + t_o^c \Leftrightarrow p_o = (R_o^c)^{-1}(p_c - t_o^c) \\ = R_c^o p_c - R_c^o t_o^c \qquad (31)$$

$$\begin{aligned} p_w &= R_o^w p_o + t_o^w = R_o^w R_c^o (p_c - t_o^c) + t_o^w \\ &= R_o^w R_c^o (R_v^c p_v + t_v^c - t_o^c) + t_o^w \\ &= R_o^w R_c^o R_v^c p_v + R_o^w R_c^o (t_v^c - t_o^c) + t_o^w \end{aligned} \qquad (32)$$

For the origin O of the camera coordinate system, there is

$0_c = R_v^c p_v + t_v^c$. Substituting it into the equation (32) can get its coordinates in the world coordinate system as:

$$o_w = -R_o^w R_c^o t_o^c + t_o^w \qquad (33)$$

For the origin of the AGV coordinate system, its coordinate in the world coordinate system is $R_o^w R_c^o (t_v^c - t_o^c) + t_o^w$, and the rotation matrix of the AGV coordinate system relative to the world coordinate system is $R_v^w = R_o^w R_c^o R_v^c$. To calculate the pose of the AGV relative to the QR code, set $R_o^w$ as the identity matrix and $t_o^w$ as the zero vector in the above formula.

## 5.2. Attitude angle calculation

### 5.2.1. Definition of attitude angle and its relationship with rotation matrix

The following describes the relationship between the rotation matrix and the attitude angle. According to the general habit, a fixed coordinate system is established on the vehicle body (machine body/hull), the positive direction of the $z$ axis is pointed to the sky, the positive direction of the $y$ axis is pointed forward, and the positive direction of the $x$ axis is pointed to the right. At the beginning, the fixed-link coordinate system coincides with the world coordinate system, and when it moves, there is a rotation, resulting in attitude angle. There are three attitude angles, namely the heading angle $\psi$, the pitch angle $\theta$ and the roll angle $\gamma$. Corresponding to the angle between the horizontal plane projection of the AGV and the $y$ axis of the world coordinate system, the angle between the $y$ axis of the AGV and the $xy$ plane of the world coordinate system, and the angle between the AGV $x$ axis and the $xy$ plane of the world coordinate system. These angles can be easily obtained through actual measurement.

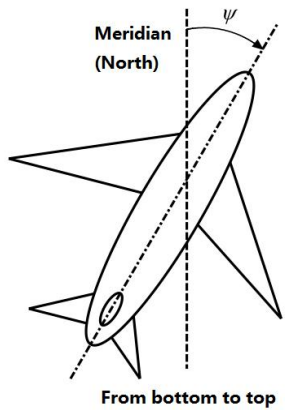

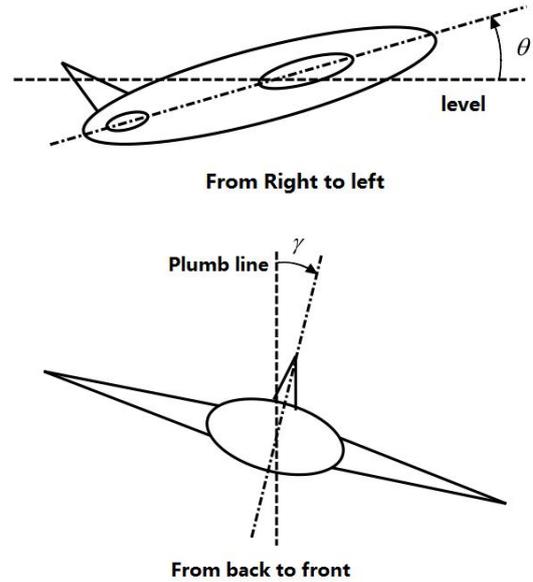

**Figure 2:** Heading angle, pitch angle and roll angle

The attitude angle can be achieved by first rotating the AGV around its own axis, then rotating around its own axis, and then rotating around its own axis. There are three issues to note here:

The attitude angle can be achieved by turning angle $\psi$ around its $z$ axis, then turning angle $\theta$ around its $x$ axis, and then turning angle $\gamma$ around its $y$ axis. There are three issues to note here:

1. The rotation sequence cannot be disrupted, otherwise the three rotation angles cannot correspond to the three attitude angles.

2. The axes in the rotation process refer to the coordinate axis of the fixed coordinate system of the AGV. After the first rotation, the direction of the AGV's $x$ axis and $y$ axis changes. After the second rotation, the AGV's $y$ axis and $z$ axis The direction changes. After the third rotation, the $x$ axis and $z$ axis directions of the AGV change;

3. From the positive direction of the rotation axis to the other two coordinate axis planes, the angle of counterclockwise rotation is positive, and clockwise is negative.

The form of the rotation matrix after the above three rotations: Set coordinate system A to coincide with coordinate system B initially, and then perform three rotations. In order to describe the facts correctly, you can select any point in the coordinate system B, and calculate the coordinates of the point in the coordinate system A after three rotations.

The rotation matrix of $\psi$ around the axis $z$ is:

$$T_z(\psi) = \begin{pmatrix} cos\psi & -sin\psi & 0 \\ sin\psi & cos\psi & 0 \\ 0 & 0 & 1 \end{pmatrix} \quad (34)$$

The rotation matrix of $\theta$ around the axis $x$ is:

$$T_x(\theta) = \begin{pmatrix} 1 & 0 & 0 \\ 0 & cos\theta & -sin\theta \\ 0 & sin\theta & cos\theta \end{pmatrix} \quad (35)$$

The rotation matrix of $\gamma$ around the axis $y$ is:

$$T_y(\gamma) = \begin{pmatrix} cos\gamma & 0 & sin\gamma \\ 0 & 1 & 0 \\ -sin\gamma & 0 & cos\gamma \end{pmatrix} \quad (36)$$

For any point P in space, set its coordinate in coordinate system B as $p_b$, and initially, its coordinate in coordinate system A as $p_b$, and set its coordinates after three successive rotations as $p_z$、$p_{zx}$、$p_{zxy} \equiv p_a$ .One can obtain:

$p_z$、$p_{zx}$、$p_{zxy} \equiv p_a$ .

$$\begin{aligned} p_b &= T_z(\psi)p_z = T_z(\psi)T_x(\theta)p_{zx} \\ &= T_z(\psi)T_x(\theta)T_y(\gamma)p_a \equiv R_{\psi\theta\gamma}p_a \end{aligned} \quad (37)$$

where $R_{\psi\theta\gamma} = \begin{pmatrix} c_\psi c_\gamma - s_\psi s_\theta s_\gamma & -s_\psi c_\theta & c_\psi s_\gamma + s_\psi c_\gamma s_\theta \\ s_\psi c_\gamma + c_\psi s_\theta s_\gamma & c_\psi c_\theta & s_\psi s_\gamma - c_\psi s_\theta c_\gamma \\ -c_\theta s_\gamma & s_\theta & c_\theta c_\gamma \end{pmatrix}$ (38)

where $c_x = cosx, s_x = sinx$.

### 5.2.2. When the pitch angle is too large

From formula (38), one can obtain:

$$\theta = arcsinr_{32} \quad (39)$$

Since the arctangent function is symmetric about $\frac{\pi}{2}$ in the interval [0, $\pi$], and symmetric about $-\frac{\pi}{2}$ in the interval [-$\pi$, 0], $\theta$ is limited to the interval $[-\frac{\pi}{2}, \frac{\pi}{2}]$, which naturally leads to a problem: if the elevation angle exceeds $\pm\frac{\pi}{2}$, what should be done? This is equivalent to the current rotation turning backward and flipping. If $\pi \geq \theta > \frac{\pi}{2}$, then the rotation $R_{\psi\theta\gamma}$ is equivalent to the rotation $R_{\psi'\theta'\gamma'}$ of $\psi' = \psi + \pi$、$\theta' = \pi-\theta$、$\gamma' = \gamma + \pi$ .If $-\pi \leq \theta < -\frac{\pi}{2}$, then the rotation $R_{\psi\theta\gamma}$ is equivalent to the rotation $R_{\psi'\theta'\gamma'}$ of $\psi' = \psi + \pi$、$\theta' = -\pi-\theta$、$\gamma' = \gamma + \pi$ .And in both cases $R_{\psi\theta\gamma} = R_{\psi'\theta'\gamma'}$.

### 5.2.3. The condition that the AGV is perpendicular to the ground

When $\theta \neq \pm\frac{\pi}{2}$ 时, one can obtain:

$$c_\psi = \frac{r_{22}}{c_\theta}, s_\psi = -\frac{r_{12}}{c_\theta}, c_\theta > 0 \quad (40)$$

$$c_\gamma = \frac{r_{33}}{c_\theta}, s_\gamma = -\frac{r_{31}}{c_\theta}, c_\theta > 0 \quad (41)$$

When $\theta = \pm\frac{\pi}{2}$, from the matrix form, it is not easy to get the values of $\psi$ and $\gamma$, then

$$R_{\psi\frac{\pi}{2}\gamma} = \begin{pmatrix} c_\psi c_\gamma - s_\psi s_\gamma & 0 & c_\psi s_\gamma + s_\psi c_\gamma \\ s_\psi c_\gamma + c_\psi s_\gamma & 0 & s_\psi s_\gamma - c_\psi c_\gamma \\ 0 & 1 & 0 \end{pmatrix} \quad (42)$$

$$R_{\psi-\frac{\pi}{2}\gamma} = \begin{pmatrix} c_\psi c_\gamma + s_\psi s_\gamma & 0 & c_\psi s_\gamma - s_\psi c_\gamma \\ s_\psi c_\gamma - c_\psi s_\gamma & 0 & s_\psi s_\gamma + c_\psi c_\gamma \\ 0 & -1 & 0 \end{pmatrix} \quad (43)$$

When $\theta = \frac{\pi}{2}$, for (42), one can obtain:

$$\begin{aligned} c_\psi c_\gamma - s_\psi s_\gamma &= r_{11} \\ s_\psi c_\gamma + c_\psi s_\gamma &= r_{13} \end{aligned} \quad (44)$$

Set $c_\psi = a, s_\psi = b, a^2 + b^2 = 1$, Substituting into equation (44), one can obtain:

$$\begin{aligned} c_\gamma &= ar_{11} + br_{13} \\ s_\gamma &= ar_{13} - br_{11} \end{aligned} \quad (45)$$

$R_{\psi\theta\gamma}$ is an orthogonal matrix, one can obtain $r_{11}^2 + r_{13}^2 = 1$, therefore, it can be verified $c_\gamma^2 + s_\gamma^2 = 1$. Therefore, equation (45) is an effective solution. Similarly, the value of $\gamma$ can be arbitrarily selected to obtain the corresponding $\psi$. The situation is similar for $\theta = -\frac{\pi}{2}$. Therefore, there are no array solutions for $\psi$ and $\gamma$ if $\theta = \pm\frac{\pi}{2}$. In these solutions, $\psi$ and $\gamma$ can take any value independently.

One can set $\cos\phi = r_{11}, \sin\phi = r_{13}$, according to (45) one can obtain:

$$c_\gamma = c_\psi c_\phi + s_\psi s_\phi = c_{\phi-\psi} \tag{46}$$

$$s_\gamma = c_\psi s_\phi - s_\psi c_\phi = s_{\phi-\psi} \tag{47}$$

When $\theta = \frac{\pi}{2}$, one can obtain:

$$\gamma + \psi = \phi + 2k\pi \tag{48}$$

When $\theta = -\frac{\pi}{2}$ according to (43)

$$\begin{aligned} c_\psi c_\gamma + s_\psi s_\gamma &= r_{11} \\ c_\psi s_\gamma - s_\psi c_\gamma &= r_{13} \end{aligned} \tag{49}$$

One can set $c_\psi = a, s_\psi = b, a^2 + b^2 = 1$, substituting into equation (49), one can get:

$$\begin{aligned} c_\gamma &= ar_{11} - br_{13} \\ s_\gamma &= br_{11} + ar_{13} \end{aligned} \tag{50}$$

Set $\cos\phi = r_{11}, \sin\phi = r_{13}$, according to (50) one can obtain:

$$c_\gamma = c_\psi c_\phi - s_\psi s_\phi = c_{\phi+\psi} \tag{51}$$

$$s_\gamma = c_\psi s_\phi + s_\psi c_\phi = s_{\phi+\psi} \tag{52}$$

when $\theta = -\frac{\pi}{2}$, then

$$\gamma - \psi = \phi + 2k\pi \tag{53}$$

When $\theta = \pm\frac{\pi}{2}$, the AGV is perpendicular to the ground, so the rotation of the first step is actually equivalent to that of the third step. When $\theta = \frac{\pi}{2}$, the forward rotation of the first step is equivalent to the forward rotation of the third step; When $\theta = -\frac{\pi}{2}$, the forward rotation of the first step is equivalent to the reverse rotation of the third step. In equations (48) and (53), $\phi$ is the angle of the third step when the current rotation is equivalent to the first step's rotation as zero, and both are valid when $\theta = \pm\frac{\pi}{2}$.

Therefore when $\theta = \pm\frac{\pi}{2}$, the heading angle $\psi$ and the roll angle $\phi$ are indistinguishable. According to the above discussion, at this time, one can uniformly designate the heading angle as 0 and the roll angle as $\phi$. At this time one can obtain:

$$\boldsymbol{R}_{0\frac{\pi}{2}\gamma} = \begin{pmatrix} c_\gamma & 0 & s_\gamma \\ s_\gamma & 0 & -c_\gamma \\ 0 & 1 & 0 \end{pmatrix} \tag{54}$$

$$\boldsymbol{R}_{0-\frac{\pi}{2}\gamma} = \begin{pmatrix} c_\gamma & 0 & s_\gamma \\ -s_\gamma & 0 & c_\gamma \\ 0 & -1 & 0 \end{pmatrix} \tag{55}$$

$$c_\gamma = r_{11}, s_\gamma = r_{13} \tag{56}$$

### 5.2.4. Summary

Due to the influence of precision, when $\theta = \pm\frac{\pi}{2}$, $r_{32}$ in the rotation matrix obtained by complex calculation is usually not accurate to $\pm 1$, and other elements in the same column with its peers are usually not accurate to 0.Therefore, a threshold $0 < T < 1$ should be set. When $|r_{32}| > T$, it can be considered that $\theta = \pm \frac{\pi}{2}$. The calculation of rotation angle is summarized as follows:

$$\begin{cases} \begin{cases} \theta = \arcsin(r_{32}), \theta \in (-\frac{\pi}{2}, \frac{\pi}{2}) \\ c_\psi = \frac{r_{22}}{c_\theta}, s_\psi = -\frac{r_{12}}{c_\theta}, c_\theta > 0, \quad |r_{32}| \leq T \\ c_\gamma = \frac{r_{33}}{c_\theta}, s_\gamma = -\frac{r_{31}}{c_\theta}, c_\theta > 0 \end{cases} \\ \begin{cases} \begin{cases} \theta = \frac{\pi}{2}, r_{32} > T \\ \theta = -\frac{\pi}{2}, r_{32} < -T, \quad |r_{32}| > T \end{cases} \\ \psi = 0 \\ c_\gamma = r_{11} \\ s_\gamma = r_{13} \end{cases} \end{cases} \quad (57)$$

## 6. Simulation verification

### 6.1. Comparison between manifold gradient algorithm and P4P algorithm

All the comparisons are carried out under the same camera parameters, the same pose parameters, the same feature points (in this paper, four vertices of QR image square are selected) and the same power noise conditions.

(1)Parameter group 1: the number of feature points is 4, and the coordinates of feature points are:
[-0.1333, -0.1333] [-0.1333, 0.1333] [0.1333,0.1333] [0.1333,-0.1333]，Pixel coordinates are:
[-65.1,-65.1][-65.1,143.2][143.2,143.2][143.2,65.1]，The pose is $\boldsymbol{R} = \boldsymbol{I}, \boldsymbol{t} = [0.05, 0.05, 2]^T$, and the noise is 15dB additive white Gaussian noise.
a)P4P algorithm performance:

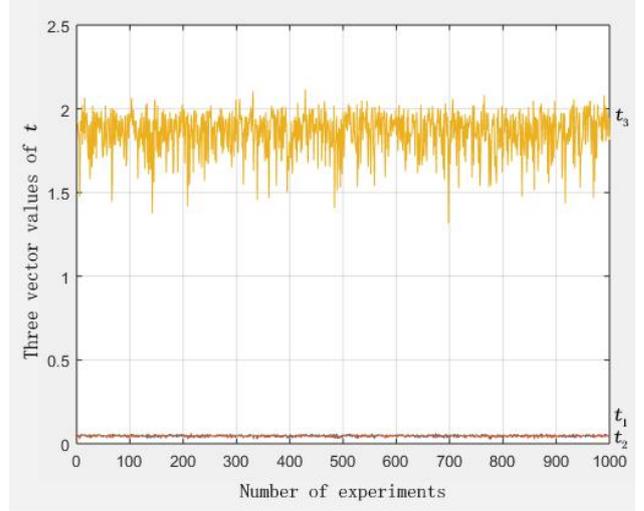

**Figure 3:** P4P algorithm performance: the horizontal axis represents the number of simulation experiments,the vertical axis represents the estimated mean value of $\boldsymbol{t}$ ,including three components $t_1$, $t_2$ and $t_3$.The estimated mean value of $\boldsymbol{t}$ obtained after 1000 calculations is [0.0468,0.0468,1.8696], and the variance is [2.6e-5,2.77e-5,1.3e-2].

b) Performance of manifold gradient optimization algorithm：

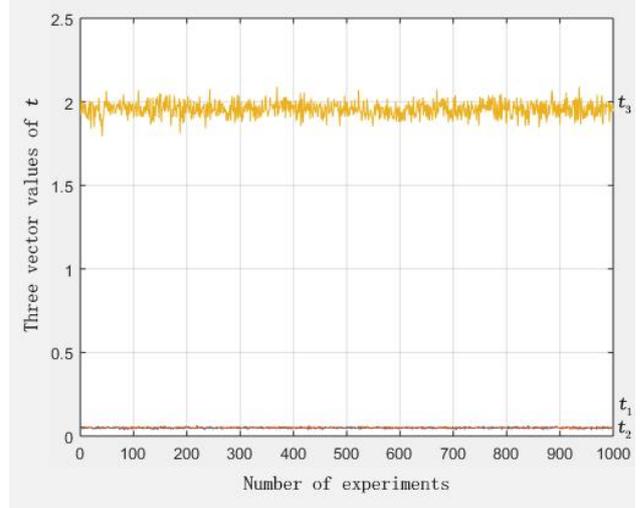

**Figure 4:** Performance of manifold gradient optimization algorithm ： the horizontal axis represents the number of simulation experiments,the vertical axis represents the estimated mean value of $\boldsymbol{t}$ ,including three components $t_1$, $t_2$ and $t_3$.The estimated mean value of $\boldsymbol{t}$ obtained after 1000 calculations is [0.0494,0.0495,1.9555], and the variance is [1.6e-5,1.57e-5,1.9e-3].

(2)Parameter group 2:the number of feature points is 4, and the coordinates of feature points are:
[-0.2667, -0.2667] [-0.2667, 0.2667] [0.2667, 0.2667] [0.2667,

-0.2667],Pixel coordinates are:
[-169.3,-169.3][-169.3,247.4][247.4,247.4][247.4,-169.3]，The pose is $R = I, t = [0.05, 0.05, 2]^T$ ,and the noise is 15dB additive white Gaussian noise.

a)P4P algorithm performance:

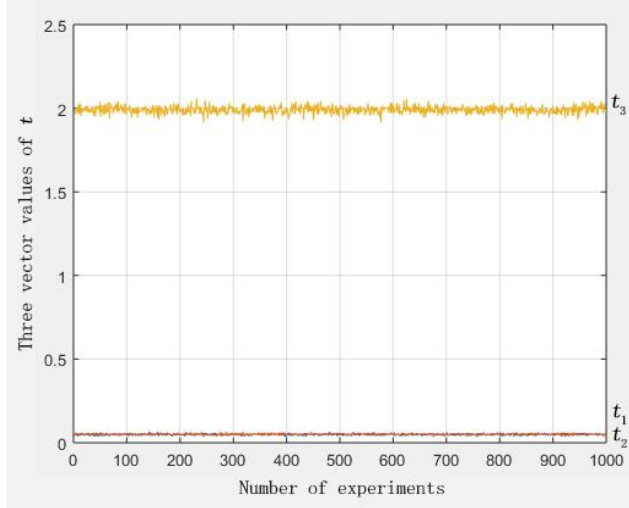

**Figure 5:** P4P algorithm performance: the horizontal axis represents the number of simulation experiments,the vertical axis represents the estimated mean value of $t$ ,including three components $t_1$, $t_2$ and $t_3$.The estimated mean value of $t$ obtained after 1000 calculations is [0.0497,0.0495,1.9885], and the variance is[2.6e-5,2.7e-5,4.1e-4].

b) Performance of manifold gradient optimization algorithm：

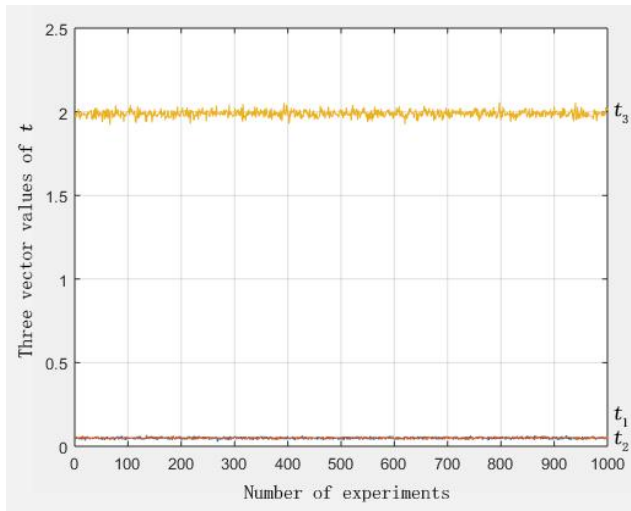

**Figure 6:** Performance of manifold gradient optimization algorithm ：the horizontal axis represents the number of simulation experiments,the vertical axis represents the estimated mean value of $t$ ,including three components $t_1$, $t_2$ and $t_3$.

The estimated mean value of $t$ obtained after 1000 calculations is [0.05,0.0499,1.9902], and the variance is [2.3e-5,2.5e-5,4.6e-4].

(3)Parameter group 3:the number of feature points is 4, and the coordinates of feature points are:

[-0.2667, -0.2667] [-0.2667, 0.2667] [0.2667, 0.2667] [0.2667, -0.2667]，Pixel coordinates are:

[-169.3,-169.3][-169.3,247.4][247.4,247.4][247.4,-169.3]，The pose is $R = I, t = [0.05, 0.05, 2]^T$ ,and the noise is 22dB additive white Gaussian noise.

a)P4P algorithm performance:

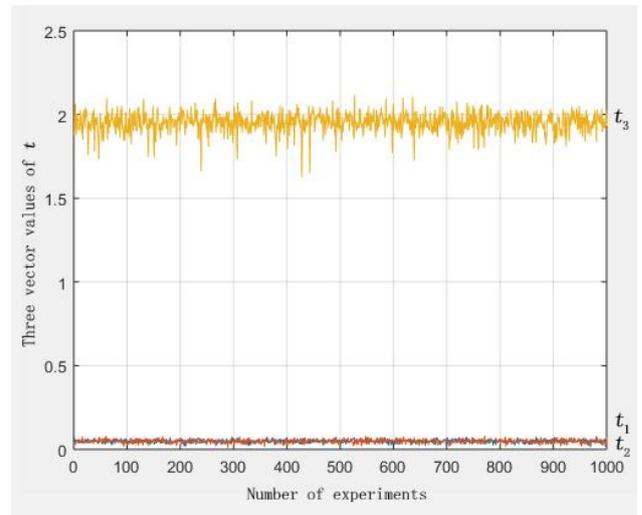

**Figure 7:** P4P algorithm performance: the horizontal axis represents the number of simulation experiments,the vertical axis represents the estimated mean value of $t$ ,including three components $t_1$, $t_2$ and $t_3$.The estimated mean value of $t$ obtained after 1000 calculations is [0.0487,0.0488,1.9506], and the variance is [1.1e-4,1.3e-5,3.7e-3].

b) Performance of manifold gradient optimization algorithm：

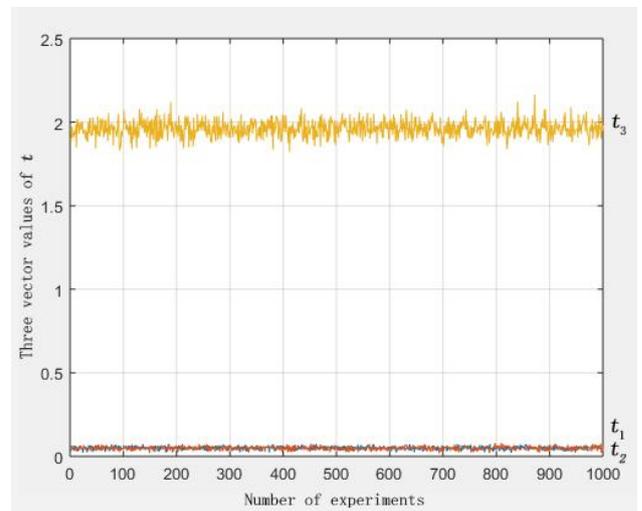

**Figure 8:** Performance of manifold gradient optimization algorithm：the horizontal axis represents the number of simulation experiments,the vertical axis represents the estimated mean value of $t$ ,including three components $t_1$, $t_2$ and $t_3$. The estimated mean value of $t$ obtained after 1000 calculations is [0.0489,0.0490,1.9509], and the variance is [1.1e-4,1.3e-5,3.3e-3].

(4)Parameter group 4:the number of feature points is 4, and the coordinates of feature points are:

[-0.2667, -0.2667] [-0.2667, 0.2667] [0.2667, 0.2667] [0.2667, -0.2667]，Pixel coordinates are:

[-169.3,-169.3][-169.3,247.4][247.4,247.4][247.4,-169.3]，The pose is

$$R = I, t = [0.05, 0.05, 2]^T.$$

Apply 15/16/17/18/19/20/21/22/23/24 dB additive white Gaussian noise respectively, and use P4P algorithm and manifold gradient optimization algorithm to calculate 1000 times respectively under each additive white Gaussian noise environment, and compare the values of $t$ components obtained under the two algorithms.

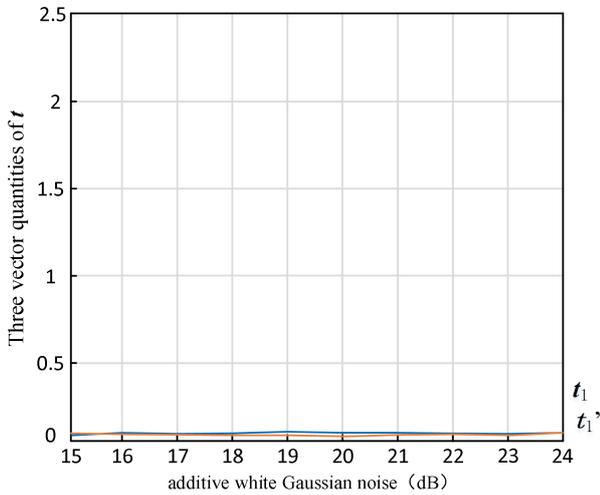

**Figure 9:** Performance comparison between P4P algorithm and manifold gradient optimization algorithm.Both algorithms are calculated 1000 times under the additive white Gaussian noise marked by the horizontal axis to obtain the first component of t value. The blue line $t_1$ represents the calculation result of P4P algorithm, and the red line $t_1$' represents the calculation result of manifold gradient optimization algorithm.

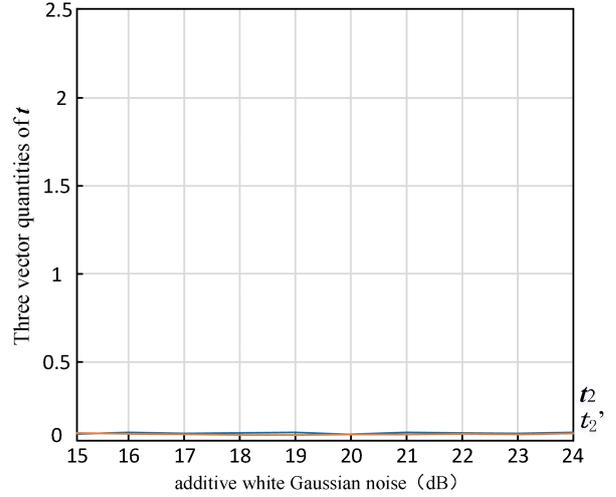

**Figure 10:** Performance comparison between P4P algorithm and manifold gradient optimization algorithm.Both algorithms are calculated 1000 times under the additive white Gaussian noise marked by the horizontal axis to obtain the first component of t value. The blue line $t_2$ represents the calculation result of P4P algorithm, and the red line $t_2$' represents the calculation result of manifold gradient optimization algorithm.

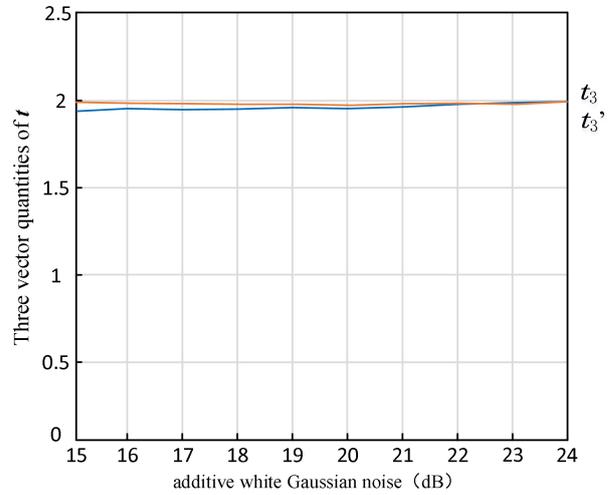

**Figure 11:** Performance comparison between P4P algorithm and manifold gradient optimization algorithm.Both algorithms are calculated 1000 times under the additive white Gaussian noise marked by the horizontal axis to obtain the first component of t value. The blue line $t_3$ represents the calculation result of P4P algorithm, and the red line $t_3$' represents the calculation result of manifold gradient optimization algorithm.

In Figure 3 and Figure 4, the noise is 15dB additive Gaussian white noise,the estimated mean value of $t$ in Figure 4 is greater than the estimated mean value of $t$ in Figure

3,and the $t$ difference is [0.0494-0.0468,0.0495-0.0468,1.9555-1.8696],that is [0.0026,0.0027,0.0859],the variance in Figure 4 is less than the variance in Figure 3,that is 1.6e-5<2.6e-5,1.57e-5<2.77e-5,1.9e-3<1.3e-2.In Figure 5 and Figure 6, the noise is 15dB additive Gaussian white noise,the estimated mean value of $t$ in Figure 6 is greater than the estimated mean value of $t$ in Figure 5,and the $t$ difference is [0.05-0.0497,0.0499-0.0495,1.9902-1.9885],that is [0.0003,0.0004,0.0017],the variance in Figure 6 is less than the variance in Figure 5,that is 2.3e-5<2.6e-5,2.5e-5<2.7e-5,4.6e-4>4.1e-4.In Figure 7 and Figure 8, the noise is 22dB additive Gaussian white noise,the estimated mean value of $t$ in Figure 8 is approximately equal to the estimated mean value of $t$ in Figure 7, 0.0489≈0.0487,0.0490≈0.0488,1.9509≈1.9506,the variance in Figure 4 is approximately equal to the variance in Figure 3,that is 1.1e-4=1.1e-4,1.3e-5<1.3e-5,3.3e-3≈3.7e-3.From the above simulation comparison, one can see that the performance of the manifold gradient optimization algorithm proposed in this paper is much better than that of the P4P analytic algorithm when the signal-to-noise ratio is small.

In Figures 9, 10 and 11,one can see that with the increase of the signal-to-noise ratio, the performance of the P4P analytic algorithm approaches that of the manifold gradient optimization algorithm.

## 6.2. Comparison of algorithm performance when the number of feature points changes

(1)Parameter group 1:the number of feature points is 4 (in this paper, four vertices of QR image square are selected) , and the coordinates of feature points are:
[-0.2667, -0.2667] [-0.2667, 0.2667] [0.2667, 0.2667] [0.2667, -0.2667]，Pixel coordinates are:
[-169.3,-169.3][-169.3,247.4][247.4,247.4][247.4,-169.3]，The pose is
$$R = I, t = [0.05, 0.05, 4]^T$$ ,and the noise is 22dB additive white Gaussian noise.

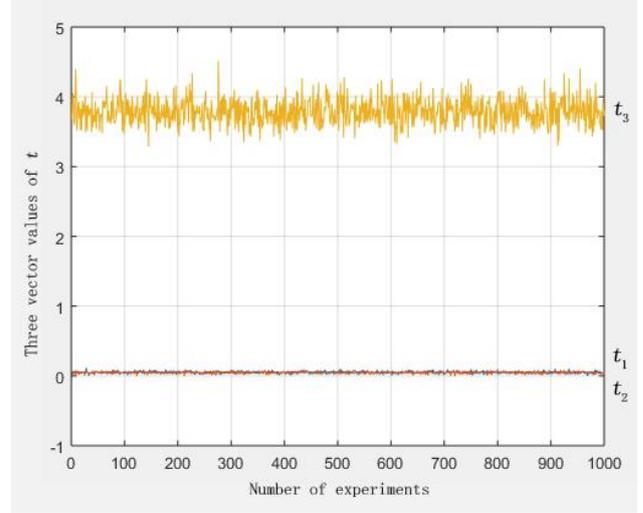

**Figure 12:** Performance of manifold gradient optimization algorithm：the horizontal axis represents the number of simulation experiments,the vertical axis represents the estimated mean value of $t$ ,including three components $t_1$, $t_2$ and $t_3$.The estimated mean value of $t$ obtained after 1000 calculations is [0.0504,0.0492,3.7869], and the variance is[2.4e-4,2.5e-5,3.3e-2].

the number of feature points is 8(these are the four vertices of the square of the QR code picture and the midpoints of the four sides of the square.), the other parameters are the same.

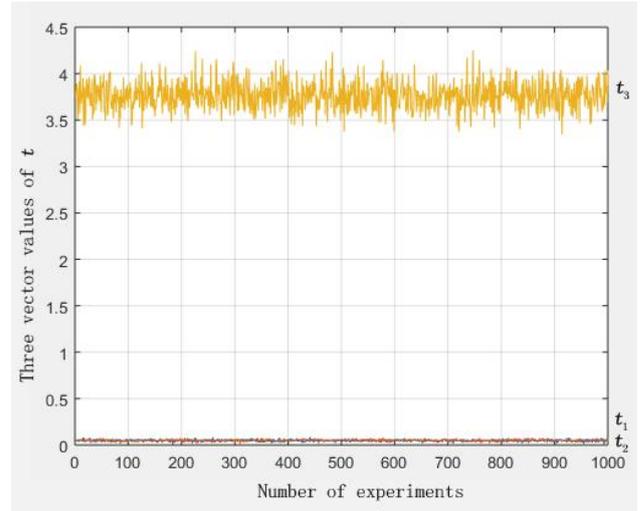

**Figure 13:** Performance of manifold gradient optimization algorithm：the horizontal axis represents the number of simulation experiments,the vertical axis represents the estimated mean value of $t$ ,including three components $t_1$, $t_2$ and $t_3$.The estimated mean value of $t$ obtained after 1000 calculations is[0.0492,0.0491,3.7646], and the variance is [1.2e-4,1.2e-5,2.2e-2]

(2)Parameter group 2:the number of feature points is 4 (in this paper, four vertices of QR image square are selected) , and the coordinates of feature points are:

[-0.1333, -0.1333] [-0.1333, 0.1333] [0.1333, 0.1333] [0.1333, -0.1333]，Pixel coordinates are:

[-32.6,-32.6][-32.6,71.6][71.6,71.6][71.6,-32.6]

，The pose is $R = I, t = [0.05, 0.05, 4]^T$ ,and the noise is 22dB additive white Gaussian noise.

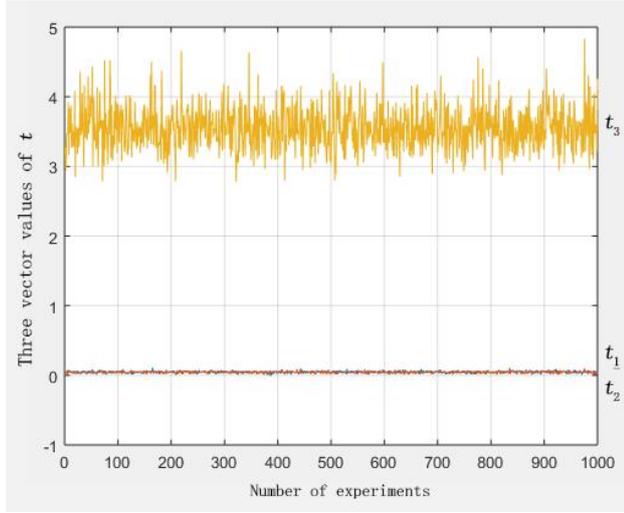

**Figure 14:** Performance of manifold gradient optimization algorithm：the horizontal axis represents the number of simulation experiments,the vertical axis represents the estimated mean value of $t$ ,including three components $t_1$, $t_2$ and $t_3$.The estimated mean value of $t$ obtained after 1000 calculations is [0.0459,0.0453,3.5480], and the variance is [2.1e-4,2.2e-4,9.7e-2] .

the number of feature points is 8(these are the four vertices of the square of the QR code picture and the midpoints of the four sides of the square.), the other parameters are the same.

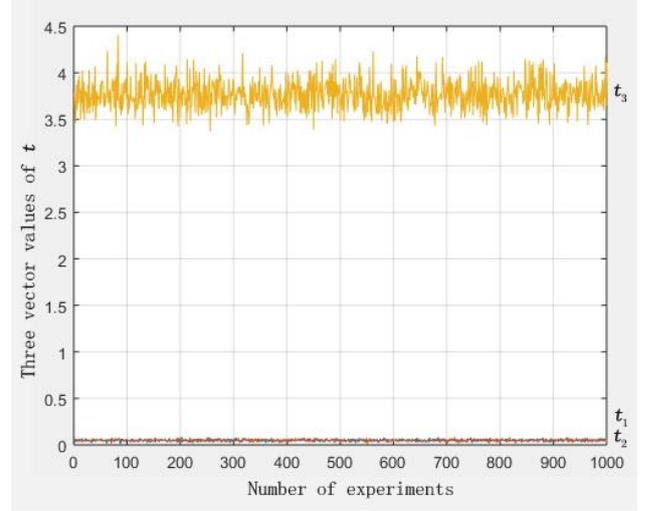

**Figure 15:** Performance of manifold gradient optimization algorithm：the horizontal axis represents the number of simulation experiments,the vertical axis represents the estimated mean value of $t$ ,including three components $t_1$, $t_2$ and $t_3$.The estimated mean value of $t$ obtained after 1000 calculations is [0.0488,0.0488,3.7663], and the variance is [1.3e-4,1.3e-5,2.4e-2].

In Figure 12 ,Figure 13,Figure 14 and Figure 15,the noise is the same,that is 22dB additive white Gaussian noise.In Figure 12 the number of feature points is 4,in Figure 13 the number of feature points is 8,in Figure 14 the number of feature points is 4,in Figure 15 the number of feature points is 8.The variance in Figure 13 is less than the variance in Figure 12,that is 1.2e-4<2.4e-4,1.2e-5<2.5e-5,2.2e-2<3.3e-2, the number of feature points in Figure 13 is more than that in Figure 12,the data of $t$ in Figure 13 fluctuates less, so the performance is better.The variance in Figure 15 is less than the variance in Figure 14,that is 1.3e-4<2.1e-4,1.3e-5<2.2e-4,2.4e-2<9.7e-2,the number of feature points in Figure 15 is more than that in Figure 14,the data of $t$ in Figure 15 fluctuates less, so the performance is better.

From the simulation comparison in Section 6.1, one can see that the performance of the manifold gradient optimization algorithm proposed in this paper is much better than that of the P4P analytic algorithm when the signal-to-noise ratio is small.With the increase of the signal-to-noise ratio, the performance of the P4P analytic algorithm approaches that of the manifold gradient optimization algorithm.

From the simulation comparison in Section 6.2,one can see

that when the noise is same, the performance of manifold gradient optimization algorithm is better when there are more feature points.

## 7. experimental verification

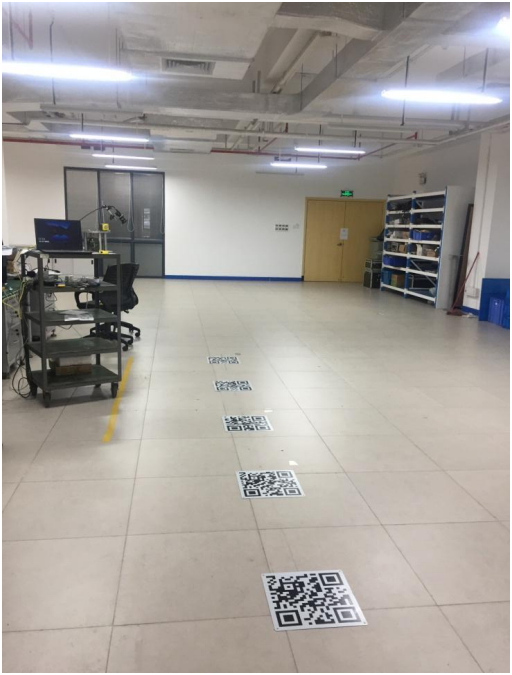

**Figure 16:** Realistic view of QR code test environment

The two-dimensional code test environment scene camera is installed on a movable shelf car through the camera frame to simulate the two-dimensional code positioning in the moving state. The camera frame can adjust the camera installation angle.

Because the information stored in the QR code is the coordinates of the plane coordinate system, it is necessary to define a plane coordinate system on the ground of the laboratory.

After the test environment is built and the test camera also meets the vision and pixel accuracy requirements of the algorithm, change the size of the QR code, the height and tilt angle of the camera, and the location solution module calculates the camera position.

One experimental method is to rotate the camera around the center of a single two-dimensional code, the camera height is 1500mm, and the horizontal distance from the center of the two-dimensional code is 1000mm. The camera rotates around the center point of the two-dimensional code every 2°, and calculates the estimation of the origin coordinates of the two-dimensional code through P4P and manifold gradient optimization algorithm Value and camera rotation angle. The final statistics of 1000 times show that the camera coordinate position error calculated by this algorithm is within ±2mm, and the rotation angle error is within ±0.5°

Another experimental method is to calculate the positioning accuracy of AGV trolley in motion. The QR code is pasted at a distance of 1000mm, and the AGV motion speed is 0.5,1.0,1.5,2.0,2.5m/s respectively. After 100 experiments at each speed, the camera coordinate position error is within ±2mm, and the rotation angle error is within ±0.5°

The positioning accuracy of this algorithm is compared with that of relevant documents. For example, the technical scheme of combining laser and two-dimensional code technology is adopted in document[29]. The feasibility of this guidance scheme is verified by simulation, which makes up for the deficiency of single guidance method, reduces the guidance error, and the positioning accuracy can be maintained at 10-15mm. Document [30] adopts the navigation method based on the DM (Datamatrix) two-dimensional code AGV, stores the position information in the DM code, and uses the position information stored in the DM code and the significant feature L edge of the DM code for positioning and navigation. When the test speed is 1,1.5,2,2.5m/s, the two-dimensional code position information in the AGV operation process is obtained, with the maximum angle error greater than 1° and the minimum angle error of 0.5°, The maximum position error is close to 20mm, and the minimum position error is 2mm.

The simulation and experimental results show that the algorithm has high positioning accuracy and stability.

## 8. Measures and suggestions for further improving positioning accuracy

The QR code navigation and positioning algorithm described in this paper can be divided into two parts: firstly, four feature points are selected for rough estimation, and then iterative optimization is performed to improve the accuracy. The latter part does not limit the number of feature points.

The limitation of this algorithm is that the four selected feature points should be on the same plane (if they are not on the same plane, there is the possibility of multiple solutions). If the feature points are not coplanar, more feature points need to be selected to complete the previous work. P4P algorithm provides the estimated initial value, so the accuracy requirement is not too high, and four feature points are enough.

This algorithm depends on the correct positioning of QR code, but the QR code currently used is not robust enough in terms of code type. In the processing method, it is required that all three positioning marks should be identified, and the positioning of pixel coordinates of feature points depends on edge recognition.

In the future optimization, the research need to improve the code pattern to improve the decoding accuracy, improve the pixel coordinate estimation accuracy of feature points and add more feature points. Increasing the number of feature points has three advantages: 1. Improving the estimation accuracy of iterative optimization; 2. When some feature points cannot be extracted, the positioning processing can still be completed by the remaining points, which greatly improves the robustness; 3. It can provide convenience for feature point optimization, and in the process, feature points with larger errors can be removed to improve performance.

The improvement can be carried out step by step. In the first stage, QR code can be used continuously, but circular or square feature points are added at the edge of QR; In the second stage, the comprehensive optimization is improved by combining the type of QR code.

## 9. Conclusions

This paper introduces a method of collecting precise position and pose state of AGV by information physics system applied to robot. Through the P4P algorithm to calculate the QR code system position relative to the camera coordinate system, then put the QR code coordinate position relative to the camera coordinate system transformation for AGV body relative to the world coordinate system of the position and posture of AGV can be precise location information, in order to obtain more accurate car location information, design the manifold gradient optimization algorithm, After the rotation matrix is calculated by the P4P algorithm, the rotation matrix is put into the gradient optimization algorithm for iterative optimization. Simulation results show that the performance of P4P algorithm optimized by manifold gradient optimization algorithm is much better than that of P4P analytical algorithm, and the positioning accuracy is effectively improved. At the same time, this paper puts forward the precautions of using the algorithm described in this paper and suggestions for further improving the positioning accuracy.

One of the measures is to obtain higher positioning accuracy by increasing the number of feature points on the two-dimensional code for calculation. The more the number of feature points, the higher the positioning accuracy of P4P algorithm is verified by simulation. Manifold gradient optimization algorithm can also improve the performance of P4P algorithm.